\documentclass[a4paper]{article}
\usepackage{iwslt18,amssymb,amsmath,epsfig}
\def\reg{{\rm\ooalign{\hfil
     \raise.07ex\hbox{\scriptsize R}\hfil\crcr\mathhexbox20D}}}

\usepackage{url}
\usepackage{xspace}
\usepackage{xcolor}

\newcommand\BLEU{\textsc{Bleu}\xspace}
\newcommand\TER{\textsc{Ter}\xspace}

\usepackage{pmboxdraw}
\pmboxdrawbox{A}

\definecolor{armygreen}{rgb}{0.0, 0.42, 0.24}

\title{Novel Applications of Factored Neural Machine Translation }

\makeatletter
\def\name#1{\gdef\@name{#1\\}}
\makeatother
\name{{\em Patrick Wilken, Evgeny Matusov}}
\address{
{AppTek}\\
{Aachen, Germany}\\
{pwilken@apptek.com, ematusov@apptek.com}
}
\begin{document}
\maketitle
\begin{abstract}
In this work, we explore the usefulness of target factors in neural machine translation (NMT) beyond their original purpose of
predicting word lemmas and their inflections, as proposed by Garc\'ia-Mart\'inez et al.~\cite{garcia2016factored}.
For this, we introduce three novel applications of the factored output architecture:
In the first one, we use a factor to explicitly predict the word case separately from the target word itself. This allows for information to be shared between different casing variants of a word.
In a second task, we use a factor to predict when two consecutive subwords have to be joined, eliminating the need for 
target subword joining markers.
The third task is the prediction of special tokens of the operation sequence NMT model (OSNMT) of Stahlberg et al.~\cite{stahlberg-etal-2018-operation}. Automatic evaluation on English$\to$German and English$\to$Turkish tasks showed that integration of such auxiliary prediction tasks into NMT is at least as good as the standard NMT approach. For the OSNMT, we observed a significant improvement in BLEU over the baseline OSNMT implementation due to a reduced output sequence length that resulted from the introduction of the target factors. 
\end{abstract}

\section{Introduction}
The state-of-the-art in machine translation are encoder-decoder neural network models \cite{Bahdanau14:softalign,transformer}. The encoder network maps the source sentence into a vector representation, while the decoder network outputs target words based on the input from the encoder.
In the standard approach, the decoder generates one output per step that directly corresponds to a single target token. In this
work, we explore applications of decoder architectures with multiple outputs per decoding step. Such an architecture was introduced for factored neural machine translation (FNMT) 
\cite{garcia2016factored} to separate the prediction of word lemmas from morphological factors such as gender and number. The main motivation of the
authors was to reduce the problem of out-of-vocabulary words. While in state-of-the-art systems this problem is tackled by the use of subword units \cite{Sennrich15:subwords, kudo2018sentencepiece}, which in theory allow for an open vocabulary, we argue that FNMT still has the following potential advantages over standard NMT:
 
\begin{enumerate}
   \item 
    Factorization can be applied on top of subword splitting. This increases the effective number of subwords for a given vocabulary size, allowing for more words to be kept in their original form.
   \item 
    Factorization can be used to explicitly share information between related (sub)words.
   \item 
    FNMT can be used 
    to produce several tokens of the target sequence 
    at the same time, decreasing the decoding sequence length and increasing decoding speed.
\end{enumerate}

To demonstrate the first two advantages, we use FNMT to pool subwords that only differ in casing or in the presence of a joining marker. 
Factors are used to make the corresponding casing or joining decisions (Section~\ref{sec:applications}). The third advantage is shown for the example of neural 
operation sequences~\cite{stahlberg-etal-2018-operation}. Here, we factor out the prediction of special tokens that move the read head 
to reduce the number of decoding steps.

\section{Related Work}

The term \textit{factored translation} first appeared in the context of statistical machine translation (SMT). The Moses system~\cite{moses} included implementation of source and target factors. They were used to either consume or 
generate different sets of label sequences: full word surface forms, lemmas, part-of-speech tags, morphological tags~\cite{koehn2007factored}. The original motivation was to reduce the data sparseness and estimation problems for the phrase-level translation models and target language models. Factored models were most effectively used in SMT for translation into morphologically rich languages, as shown e.g. in the work of~\cite{bojar2007english}. 


With the rapid emergence of NMT, 
a factored neural translation model was introduced by~\cite{garcia2016factored}. In that work, the out-of-vocabulary and large vocabulary size problems are mitigated by generating lemma and morphological tags. Our work is a re-implementation of that model, but we apply it for a different cause.

Word case prediction using a count-based factored translation model was done in \cite{shen2006jhu}, as an alternative to restoring case information in post-processing using a language model~\cite{lita2003truecasing}. Another common alternative is truecased training, where however casing variants of a word are handled as distinct tokens.
In our work, we evaluate case prediction as part of a neural MT system, which allows for an unbound context of previous target words and casing decisions.

Translation using subword units is a very popular vocabulary reduction technique, first proposed for NMT by 
\cite{Sennrich15:subwords}. 
To the best of our knowledge, in previous work the information on how to restore full words was always encoded as part of subword representation (e.g. with a joining marker @@), and cases where two subwords were identical except for the presence or absence of a joining marker were not handled.


\section{Baseline NMT Architecture}\label{technical}
We used an open-source 
toolkit \cite{returnn_acl2018} based on TensorFlow \cite{tensorflow2015} to implement a recurrent neural network (RNN) based translation model with attention similar to \cite{Bahdanau14:softalign}.
The model consists of 620-dimensional embedding layers for both the source and the target words, an encoder of 4 bidirectional LSTM layers with 1000 units each, a single-layer decoder LSTM with the same number of units and an additive attention mechanism.
We augmented the attention computations using fertility feedback similar to \cite{Tu2016:Coverage}. 

In the baseline system, the output layer of the decoder consist of a single softmax that computes one target word $y$ per decoding step.


\section{Target-side Factors}

\begin{figure}[t]
    \hspace{0.7cm}
    \includegraphics[width=0.4\textwidth]{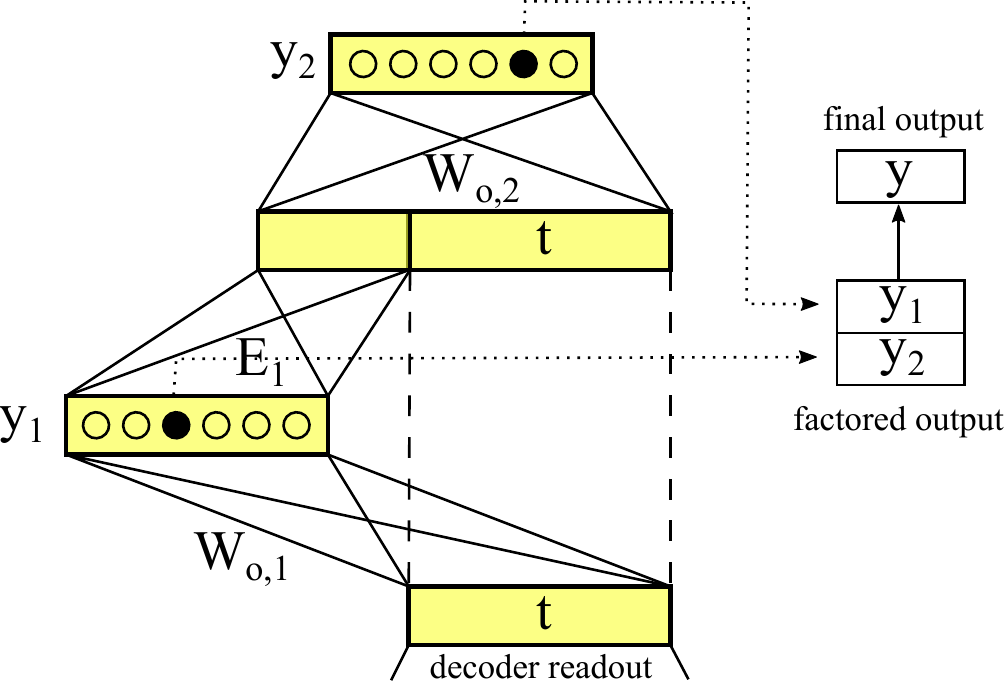}
    \caption{Structure of the factored output layer. Two outputs $y_1$ and $y_2$ are produced in every decoding step and combined in postprocessing. Crossed areas represent matrix products. Rectangles with circles represent softmax layers.}
    \label{fig:output_layer}
\end{figure}

Factored neural machine translation (FNMT) alters the decoder network such that in each decoding step two outputs $y_1$ and $y_2$ are generated instead of one. We choose an architecture similar to \cite{garcia2016factored}, see Figure~\ref{fig:output_layer}. 

The probability distribution for $y_1$ is computed directly from the readout vector $t$ of the decoder\footnote{following the notation of \cite{Bahdanau14:softalign}}:
\begin{equation}
    p(y_1|t) = \mathrm{softmax}(W_{o,1}t),
\end{equation}
where $W_{o,1}$ is the output weight matrix for the first output. We let $p(y_2)$ depend on the embedding $E_1y_1$ of the first output, which is 
called \textit{lemma dependency} in \cite{garcia2016factored}:
\begin{equation}
    \label{eq:second_factor}
    p(y_2|t, y_1) = \mathrm{softmax}(W_{o,2}\,[t; E_1y_1]\,).
\end{equation}
Here, $[\cdot]$ denotes concatenation. The hidden state $s$ of the decoder LSTM is 
the concatenation of the embeddings of both previous outputs $y_1'$ and $y_2'$:
\begin{equation}
    s = f(s', [E_1y_1', E_2y_2'], c),
\end{equation}
where $s'$ is the previous state and $c$ the context vector computed by the attention mechanism. $E_1$ and $E_2$ 
have the same embedding size as the baseline.

In training, we optimize the sum of the cross-entropies of both factors. For beam search, we first calculate the softmax of the first factor $y_1$, which results in an intermediate beam of the $n$-best hypotheses after adding scores for $y_1$. After that, for all hypotheses in the beam, the embedding of $y_1$ and the readout $t$ from which $y_1$ originated are fed into the softmax for the second factor $y_2$. Here, scores for $y_2$ according to Equation \ref{eq:second_factor} are added (in log-space). The final beam therefore contains hypotheses that were expanded by both $y_1$ and $y_2$ in the current decoding step.

Compared to the standard output layer, the factored output layer requires the computation of an additional softmax and logic to map the indices of the readout vectors $t$ of the incoming beam into the intermediate beam.
In practice however, this does not lead to an increase in decoding time, as explained in Section \ref{subsec:Speed}.

Note, that this factored output implementation can be generalized to more than two factors for more complex applications. Also, it is independent of the underlying network and can therefore be used on top of other architectures such as the Transformer~\cite{transformer}.

\begin{figure*}[t]
    \centering
    \includegraphics[width=0.8\textwidth]{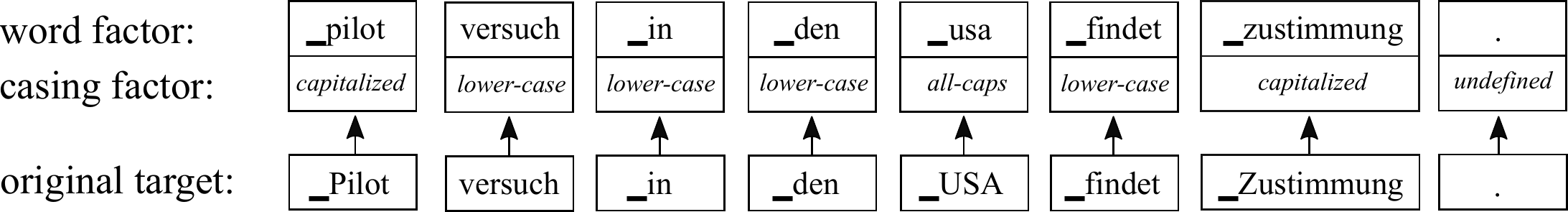}
    \caption{Factorization of casing information. The original target words are converted into a lower-case target and a casing factor. The original target can easily be restored from the two factors.}
    \label{fig:casing_example}
\end{figure*}
\begin{figure*}[t]
    \centering
    \includegraphics[width=0.8\textwidth]{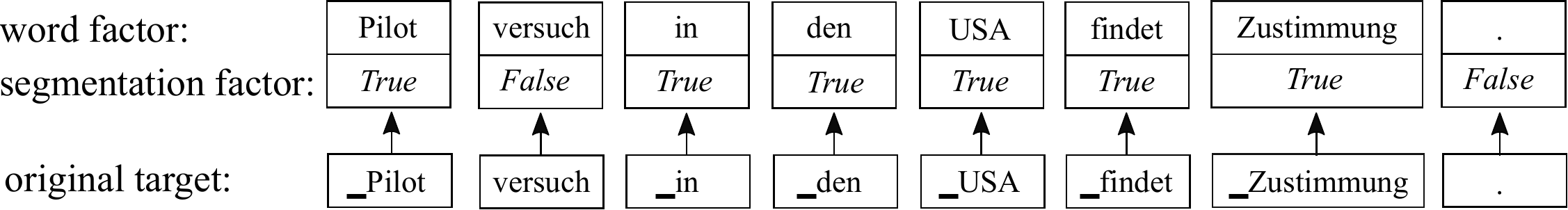}
    \caption{Factorization of subword merges. In the original target the Sentencepiece marker {\tiny{\pmboxdrawuni{2581}}} is used to mark separation between words. In the factored target a binary flag is used instead.}
    \label{fig:subword_merge_example}
\end{figure*}

\section{Applications of the Factored Model}\label{sec:applications}

\subsection{Word case prediction}\label{subsec:case_prediction}
As a first application, we use the factored output architecture to split the prediction of a target \mbox{(sub)word} into the prediction of its lower-cased form ($y_1$) and
a casing class ($y_2$). Such a factorization is illustrated in Figure \ref{fig:casing_example}. 
The model makes no distinction between different casing variants
of the same word w.r.t. the first factor $y_1$. This is 
helpful e.g. in the case
of German verb nominalization, which entails capitalization. For example, there is a direct correspondence between the words \texttt{treffen} (meet) and \texttt{Treffen} (meeting). In a truecased representation 
this connection would be lost, as these words would be represented as two distinct tokens. Another advantage of this factorization 
is that the vocabulary for the factor $y_1$ is lower-case only, and thus its size is reduced significantly. 

We use four classes for the casing factor: \textit{lower-case}, \textit{capitalized}, \textit{all-caps}, and \textit{undefined}. Each original 
target word $y$ of the target sentence is assigned to one of the classes. To fall into \textit{lower-case} or \textit{all-caps}, all
alphabetical characters in a word have to be lower- or upper-case, respectively.
For \textit{capitalized}, the first character has to be upper-case and the remaining ones lower-case.
Tokens without any letters, as well as mixed-cased words other than \textit{capitalized}, are assigned to \textit{undefined}.
In decoding, the predicted casing is applied to each corresponding word as a postprocessing step.
For simplicity, in case of the \textit{undefined} class we keep the word as is. This leads to errors for mixed cased subwords, which however make up less than $0.04\%$ of the subwords in the test sets.

\subsection{Subword segmentation}
Another application of FNMT 
is the prediction of 
subword merges. Instead of encoding whether a subword is to be merged with adjacent subwords by using a special marker as part of the subword itself, we predict merges explicitly via the second output $y_2$ of the network, as shown in Figure~\ref{fig:subword_merge_example}. The motivation behind this is a better handling of compound words. Intuitively, the representation of a compound word should be a function of its compounds. However, with traditional subword methods, distinct tokens are used for the compound, depending on whether it is part of a word or stands alone. In the example in Figure~\ref{fig:subword_merge_example}, we have the German compound word \texttt{Pilotversuch} (pilot trial). In traditional subword vocabularies, the two tokens (\texttt{\_Pilot}, \texttt{Pilot}) would be separate subword tokens, and the fact that they are related would be only learned by the network if they are both frequent enough in the training data. 
In the proposed FNMT approach, the same token $y_1$ is used instead. 

To create the two training targets $y_1$ and $y_2$, we first create a standard subword representation of the target text. 
From each subword $y$ of the target sentence we then extract a binary flag $y_2$, telling whether the subword is to be separated from its left 
neighbour or not, based on the joining markers. The first output factor $y_1$ is created by removing the joining marker from $y$ if present. 

In decoding, we insert a space left of only those tokens for which $y_2$ is predicted as \textit{True}.


\section{Factored Operation Sequence NMT}\label{osnmt}
Operation sequence neural machine translation, as proposed by \cite{stahlberg-etal-2018-operation}, uses a special representation of the target sentence from which it is possible to
extract an alignment link for each target word to one of the source words. This alignment information is potentially useful for a number of interesting applications, such as integrating external lexical rules into NMT or copying annotations, such as formatting, from source to target \cite{arthur-etal-2016-incorporating, zenkel2019adding}.

In OSNMT, the source words are translated monotonically. To encode reordering, the target vocabulary is extended by tokens to set markers in the sequence (\texttt{SET\_MARKER}) and to jump between these markers (\texttt{JMP\_FWD} and \texttt{JMP\_BWD}).
The alignment link is given by a virtual read head, which traverses the source sentence from left to right and is moved by an additional \texttt{SRC\_POP} token. It is inserted into the sequence when all target words aligned to the source word under the current read head have been produced.

In a full operation sequence the \texttt{SRC\_POP} token appears once for every source word. Therefore its length is at least the sum of the number of source 
and target words, which 
means high computational costs and, possibly, worse translation quality because of long range dependencies. We therefore propose to factor out the generation of \texttt{SRC\_POP} tokens. For this, we predict all tokens except \texttt{SRC\_POP} via the first output $y_1$ and let the second output $y_2$ decide how many \texttt{SRC\_POP} tokens are to be inserted before $y_1$, i.e. how many source positions to step forward. Consequently, the target vocabulary for $y_2$ consists
of integers ranging from $0$ up to a maximum of $V_\mathtt{SRC\_POP}$.

For training, we first create the operation sequence using the original algorithm of \cite{stahlberg-etal-2018-operation} and then determine $y_1$ and $y_2$ as described above. In decoding, the \texttt{SRC\_POP} tokens are inserted according to $y_2$ as a postprocessing step before compiling the
operation sequence to get the plain target sentence and the alignment.

\section{Experimental Results}
We performed experiments on the WMT 2019 {English$\to$German} and the WMT 2018 {English$\to$Turkish} news translation tasks \cite{bojar2018findings, barrault2019findings}.
For {En$\to$De}, we used all parallel corpora provided by WMT.
For {En$\to$Tr}, we used 
the provided SETIMES2 corpus \cite{TIEDEMANN12.463}, as well as 
additional in-house data.
We applied a number of heuristical data filtering rules, including: Only sentence pairs where both source and target side have more than 2 characters but less than 80 words are kept. Also, we require the number of source and target words to differ by a factor of 4 at most. On top of that, we used the FastText based language identification~\cite{joulin2016bag} 
to keep only those sentence pairs where both source and target language are assigned a confidence of at least 40\%.
Finally, we filtered out sentence pairs that have an 8-gram overlap with any sentence in the test data.
For En$\rightarrow$De, this resulted in a corpus of 23M lines and 456M running English words. 
For En$\rightarrow$Tr we have 36M lines and 313M words.

In preprocessing, we converted the English source side to lower case and applied frequency-based truecasing to the target side. Both sides were encoded separately into subwords using the unigram model of 
the Sentencepiece toolkit \cite{kudo2018sentencepiece}. A 
vocabulary size of 50K was used. 
For the case prediction experiments, we trained and applied the target subword model on lower-cased text (while still providing the original casing information to train the second factor). 

For the OSNMT experiments, we word-aligned the training data 
with the Eflomal toolkit~\cite{Ostling2016efmaral}. Unlike \cite{stahlberg-etal-2018-operation}, we did not convert the alignment to subword level. Instead we created the operation sequence on the 
word level\footnote{using the script~https://github.com/fstahlberg/ucam-scripts/blob/master/\linebreak t2t/align2osm.py} and only after that applied the subword encoding (without allowing splits of the special OSNMT tokens). This way, the number of \texttt{SRC\_POP} tokens is reduced to the number of source words instead 
of subwords, while still allowing to infer the alignments to the original source words.
Despite this, we still observe very long sequences of {\small \texttt{SRC\_POP}} tokens in the training data corresponding to unaligned source segments. We therefore exclude all sentences with more than 10 subsequent {\small \texttt{SRC\_POP}} operations from training (ca.~1\%). Accordingly, we set $V_\mathtt{SRC\_POP} = 10$ (see Section~\ref{osnmt}).

\begin{table}[t]
\begin{center}
\begin{tabular}{l|cc|cc}
 & \BLEU & \TER & \BLEU & \TER \\
 \cline{2-5}
\bf{English$\to$German} & \multicolumn{2}{c|}{\bf{newstest 2017}} & \multicolumn{2}{c}{\bf{newstest 2019}} \\
\hline 
RNN baseline & 27.6 & 54.8 & 37.4 & 46.7 \\
 + casing factor & 27.7 & 54.8 & 37.4 & 47.0 \\
 + segmentation factor & 27.8 & 54.7 & 37.4 & 46.9 \\
\hline
 OSNMT baseline & 23.2 & 59.3 & 27.3 & 55.8 \\
 + \texttt{SRC\_POP} factor & 25.0 & 57.4 & 32.4 & 50.7 \\
\hline
\hline
 \bf{English$\to$Turkish} & \multicolumn{2}{c|}{\bf{newstest 2017}} & \multicolumn{2}{c}{\bf{newstest 2018}} \\
\hline 
RNN baseline & 16.6 & 66.1 & 16.5 & 66.8 \\
+ casing factor & 16.3 & 66.5 & 16.2 & 67.3 \\
+ segmentation factor & 16.7 & 66.2 & 16.5 & 66.4 \\
\hline
OSNMT baseline & 9.7 & 72.9 & 9.9 & 73.4 \\
+ \texttt{SRC\_POP} factor & 14.3 & 69.2 & 14.4 & 69.8 \\
\hline
\end{tabular}
\end{center}
\caption{Automatic evaluation results (in \%).}
\label{tab:Results}
\end{table}

To train the network, we used the Adam optimizer \cite{kingma15} with an initial learning rate of $0.001$ and 
decayed the learning rate by a factor 
of $0.9$ whenever validation set perplexity increases. For En$\to$De we used \textit{newstest2015} as validation set, for En$\to$Tr \textit{newsdev2010}.
We used a layer-wise pre-training scheme \cite{returnn_acl2018},
label smoothing of 0.1 and a softmax layer dropout of $0.3$.

\subsection{Automatic Evaluation}\label{subsec:eval}
\begin{table}[t]
\begin{center}
\begin{tabular}{l|cc|cc}
 & \BLEU & \TER & \BLEU & \TER \\
 \cline{2-5}
\bf{English$\to$German} & \multicolumn{2}{c|}{\bf{newstest 2017}} & \multicolumn{2}{c}{\bf{newstest 2019}} \\
\hline 
RNN baseline (500K)& 21.2 & 61.3 & 25.2 & 56.6 \\
 + casing factor & 21.2 & 61.2 & 26.0 & 55.7 \\
 + segmentation factor & 20.7 & 61.9 & 25.2 & 57.1 \\
\hline
 OSNMT (500K) & 16.6 & 65.1 & 19.0 & 61.8 \\
 + \texttt{SRC\_POP} factor & 18.6 & 64.4 & 23.4 & 58.7 \\
\hline
\end{tabular}
\end{center}
\caption{Simulated low-resource condition results (in \%). All systems were trained on a corpus of 500K randomly sampled  En$\to$De sentences.}
\label{tab:LowResourceResults}
\end{table}

We report translation quality results on standard WMT test sets in Table~\ref{tab:Results} using case-sensitive \BLEU \cite{bleu} and \TER \cite{ter}. For En$\to$De \textit{newstest2019}, both a second factor predicting word casing as well as a second factor predicting subword merges resulted in identical $\BLEU$
scores to the baseline and only in a minor degradation in terms of \TER. On En$\to$De \textit{newstest2017}, a tendency towards improvement can be observed for these factored systems.

The En$\to$Tr results in Table~\ref{tab:Results} confirm the findings of the En$\to$De experiments. The slightly worse performance of casing prediction may indicate that it is most effective for German, where capitalized words are much more frequent.
\begin{table}[t]
\begin{center}
\begin{tabular}{l|c}
& \bf{runtime (minutes:seconds)} \\
\hline
RNN baseline & 0:41 \\
 + casing factor & 0:42 \\
 + segmentation factor & 0:41 \\
\hline
 OSNMT baseline & 1:07 \\
 + \texttt{SRC\_POP} factor & 1:01 \\
\hline
\end{tabular}
\end{center}
\caption{Decoding speed for En$\to$De \textit{newstest2019}}
\label{tab:Speed}
\end{table}

For both language pairs, the neural operation sequence model (OSNMT)
cannot achieve comparable translation quality to standard NMT in our experiments. As discussed earlier, we believe this is in part caused by long target sequence lengths. Reducing them 
by predicting \texttt{SRC\_POP} tokens via a factor improves the translation quality dramatically on all test sets, e.g.\ $5.1\%~\BLEU$ absolute for En$\to$De \textit{newstest2019}. We hope to be able to further improve OSNMT in future work to close the gap to standard NMT.

Count-based factored translation models tend to be more effective under low-resource conditions~\cite{koehn2007factored}. Also, in the original paper \cite{stahlberg-etal-2018-operation} OSNMT is evaluated on smaller sized tasks of up to 1M sentence pairs only. For comparison, we simulate low-resource conditions on the En$\to$De task by randomly sampling a subset of 500K sentence pairs from the training corpus. To prevent overfitting, we used only 2 encoder layers and label smoothing of 0.2 for these experiments. In addition, we optimized the size of the subword vocabulary to 10K. As can be seen in Table \ref{tab:LowResourceResults}, we observe mixed results in this setting. Casing prediction via a factor improves over the baseline by $0.8\%~\BLEU$ on \textit{newstest2019}. Factorized subword merging harms translation quality on \textit{newstest2017}. For OSNMT, we again see a weak performance of the baseline, but a big improvement due to factoring out \texttt{SRC\_POP} tokens.

\subsection{Decoding Speed}\label{subsec:Speed}
\begin{table*}[th]
\begin{center}
\footnotesize
\setlength\tabcolsep{3pt}
\def\arraystretch{1.3}
\begin{tabular}{l l}
\hline
\textbf{source:} & \texttt{After the official part, the illustrious gathering wandered to the \underline{show-jumping stadium} [...]} \\
\textbf{reference:} & \texttt{Nach dem offiziellen Teil wanderte die illustre Gesellschaft ins \underline{Springstadion} [...]}\\
\textbf{RNN baseline:} & \texttt{Nach dem offiziellen Teil wanderte das illustre Treffen in das \underline{Show-Jumping-Stadion} [...]}\\
\textbf{+ segm. factor:} & \texttt{Nach dem offiziellen Teil wanderte die illustre Versammlung zum \underline{Springstadion} [...]}\\
\hline
\textbf{source:} & \texttt{The \underline{technology legend}'s latest project [...]} \\
\textbf{reference:} & \texttt{Das aktuelle Projekt der \underline{Technologielegende} [...]}\\
\textbf{RNN baseline:} & \texttt{Das neueste Projekt der \underline{Technology Legend} [...]}\\
\textbf{+ segm. factor:} & \texttt{Das neueste Projekt der \underline{Technologie-Legende} [...]}\\
\hline
\textbf{source:} & \texttt{But \underline{denuclearization negotiations} have stalled.} \\
\textbf{reference:} & \texttt{Aber die \underline{Entnuklearisierungsverhandlungen} sind blockiert.}\\
\textbf{RNN baseline:} & \texttt{Aber die \underline{Verhandlungen \"uber die Denuklearisierung} sind ins Stocken geraten.}\\
\textbf{+ segm. factor:} & \texttt{Doch die \underline{Entnuklearisierungsverhandlungen} sind ins Stocken geraten.}\\
\hline
\textbf{source:} & \texttt{The four \underline{Baden-Wuerttemberg-class} vessels the Navy ordered back in 2007 [...]} \\
\textbf{reference:} & \texttt{Die vier im Jahr 2007 von der Marine bestellten Fregatten der \underline{Baden-W\"urttemberg-Klasse} [...]}\\
\textbf{RNN baseline:} & \texttt{Die vier \underline{baden-w\"urttembergischen} Schiffe, die die Marine im Jahr 2007 bestellt hat [...]}\\
\textbf{+ segm. factor:} & \texttt{Die vier Schiffe der \underline{Baden-W\"urttemberg-Klasse}, die die Marine im Jahr 2007 bestellt hat [...]}\\
\hline
\end{tabular}
\end{center}
\caption{Translation examples: baseline vs. segmentation factor}
\label{tab:SubwordMergeExamples}
\end{table*}

In Table \ref{tab:Speed} we compare the decoding speed of the different systems. The decodings were run on a GeForce GTX 1080 Ti with a beam size of 12 and a batch size of 3000 tokens. The runtime was observed to be constant over three runs for each of the systems.

Introducing target factors causes no noticeable slowdown in our settings. This has two reasons: First, the softmax for the second factor has a very small number of output classes, therefore it is neglectable compared to first softmax in terms of computational costs. Second, factorization reduces the vocabulary size of the first factor. For example, using the segmentation factor reduced the German vocabulary from 50K to 45K subwords.

The OSNMT system is slower than the RNN baseline due to long sequence lengths. However, factorization here increases the decoding speed, as multiple \texttt{SRC\_POP} tokens are predicted at once and simultaneously with the following token.

\subsection{Analysis}

We found casing prediction via a factor to work indistinguishably well as compared to using truecased subwords. Case-insensitive \BLEU and \TER scores for the baseline system and the one with the casing factor were found to be almost identical, e.g. $37.9 \%~\BLEU$ on En$\to$De \textit{newstest2019} for both systems. Also, when calculating \BLEU only on the casing classes of the translated words (see Section \ref{subsec:case_prediction}), the systems reach similar results ($60.9 \%$ vs. $60.7\% ~\BLEU$, respectively). Together with the similar case-sensitive scores (Table \ref{tab:Results}) this indicates that the systems neither differ significantly in the choice of words nor in their casing.

The lower-cased vocabulary of size 50K for the casing prediction experiments was able to represent a total of 134K distinct word casing variants from the training data. In low-resource settings, the 10K vocabulary represented 23K casing variants. We suspect that with enough training examples the baseline model is able to learn connections between different casing variants of a word, therefore a separate casing prediction does not improve translation quality. In low-resource settings, however, explicitly sharing information between casing variants can mitigate the data sparseness problem. This potentially explains the improvements seen in Table \ref{tab:LowResourceResults} comparing the casing factor to the baseline.

In Table \ref{tab:SubwordMergeExamples} we show that the segmentation factor indeed improves translation of compound words in many cases, especially of those not seen in training. In the first two examples, the baseline is not capable of translating the unseen compound nouns \texttt{show-jumping stadium} and \texttt{technology legend}. Instead, it resorts to copying the English words to the target side. Using the segmentation factor, the model is able to produce the reference words (in the second case using a dash for concatenation, which is valid). In the last two examples, although the baseline is able to produce reasonable translations, only the factored model is able to produce the complex unseen German compounds found in the reference. The positive effect however is not significant to automatic metric scores (Section \ref{subsec:eval}) as it affects only a small fraction of the words in the test sets.

\section{Conclusion}
We presented novel applications of factored NMT. 
We showed that word case information and joining of subword units can be predicted effectively by a target factor; this allows for a single representation of similar or even identical lexical items,
so that more full forms can be kept for a given subword vocabulary limit. Experiments on two language pairs confirm that this can be done without loss in MT quality and without an increase in decoding time. 
We also showed that our factored OSNMT implementation significantly improves the original non-factored one
by reducing the number of decoding steps. At the same time,
we showed that OSNMT on two WMT tasks exhibits significantly lower MT quality as compared to state-of-the-art NMT.
This result contradicts the original claims of \cite{stahlberg-etal-2018-operation}, who, however, ran experiments for translation into English.

In the future, we plan to utilize target factors in NMT for other auxiliary prediction problems. In particular, we are interested in a better factored representation of word and phrase alignments, that is able to provide high quality translations and alignments at the same time.

\bibliographystyle{IEEEtran}
\bibliography{translation}
\end{document}